\pdfoutput=1

\documentclass[11pt,table,usenames,dvipsnames]{article}
\usepackage[table]{xcolor}
\usepackage[]{acl}

\usepackage{times}
\usepackage{latexsym}
\usepackage{graphicx}
\usepackage[T1]{fontenc}

\usepackage[utf8]{inputenc}
\usepackage{booktabs}
\usepackage{microtype}

\usepackage{algorithm}

\usepackage{algpseudocode}

\renewcommand{\algorithmiccomment}[1]{\bgroup\hfill\small//~#1\egroup}
\usepackage{breqn}
\usepackage{graphicx}
\usepackage{amssymb}
\usepackage{amssymb}
\usepackage{pifont}
\newcommand{\cmark}{\ding{51}}%
\newcommand{\xmark}{\ding{55}}%

\usepackage{algorithm}

\usepackage{multirow}
\usepackage{url}
\usepackage{amsmath}
\usepackage[textsize=scriptsize]{todonotes}
\usepackage{booktabs}
\usepackage{comment}
\usepackage{enumitem}
\usepackage{tabu}
\usepackage{dsfont}
\usepackage{algpseudocode}
\usepackage{listings}

\usepackage{amsthm}

\newcommand{\bs}{\textsc{BS}}
\newcommand{\dbs}{\textsc{DBS}}
\newcommand{\dbsplus}{\textsc{DBS+}}
\newcommand{\topp}[1]{\textsc{Ncls}$_{#1}$}
\newcommand{\typ}[1]{\textsc{Typ}$_{#1}$}
\newcommand{\btopp}[1]{\textsc{BNcls}$_{#1}$}
\newcommand{\btyp}[1]{\textsc{BTyp}$_{#1}$}

\newcommand{\ar}{$\overline{\textsc{R}}$}
\newcommand{\ao}{$\overline{\textsc{OR}}$}
\newcommand{\ad}{$\overline{\textsc{D}}$}
\newcommand{\mtr}{\textsc{Mtr}}
\newcommand{\grm}{\textsc{Grm}}
\newcommand{\mv}{\textsc{Mv}}
\newcommand{\done}{\textsc{D}-1}
\newcommand{\dtwo}{\textsc{D}-2}
\newcommand{\dthree}{\textsc{D}-3}

\newcommand{\bfsmean}{\textsc{BKS}$_{mean}$ }
\newcommand{\bfslast}{\textsc{BKS}$_{last}$ }

\newcommand{\narrowcol}{\setlength{\tabcolsep}{4pt}}
\newcommand{\badcell}{\cellcolor{BrickRed!15}}
\newcommand{\finecell}{\cellcolor{RoyalBlue!10}}
\newcommand{\goodcell}{\cellcolor{RoyalBlue!25}}

\DeclareMathOperator*{\argmax}{arg\,max}

\newcommand{\colorr}[1]{\textcolor{Bittersweet}{#1}}
\newcommand{\colorg}[1]{\textcolor{BlueViolet}{#1}}
\newcommand{\colorb}[1]{\textcolor{OliveGreen}{#1}}

\title{Best-$k$ Search Algorithm for Neural Text Generation}

\author{Jiacheng Xu, Caiming Xiong, Silvio Savarese, Yingbo Zhou\\
  Salesforce AI Research \\
  \texttt{\{jiacheng.xu,cxiong,ssavarese,yingbo.zhou\}@salesforce.com}  \\}

\begin{document}
\maketitle

\begin{abstract}

Modern natural language generation paradigms require a good decoding strategy to obtain quality sequences out of the model. Beam search yields high-quality but low diversity outputs; stochastic approaches suffer from high variance and sometimes low quality, but the outputs tend to be more natural and creative. In this work, we propose a deterministic search algorithm balancing both quality and diversity. We first investigate the vanilla best-first search (BFS) algorithm and then propose the \emph{Best-$k$ Search} algorithm. Inspired by BFS, we greedily expand the top $k$ nodes, instead of only the first node, to boost efficiency and diversity. Upweighting recently discovered nodes accompanied by heap pruning ensures the completeness of the search procedure. Experiments on four NLG tasks, including question generation, commonsense generation, text summarization, and translation, show that best-$k$ search yields more diverse and natural outputs compared to strong baselines, while our approach maintains high text quality. The proposed algorithm is parameter-free, lightweight, efficient, and easy to use\footnote{Code, implementation, and visualization will be available. }.
\end{abstract}

\section{Introduction}
The recent success of large-scale pre-trained language models (\citet{devlin-etal-2019-bert,raffel2020exploring,brown2020language,nijkamp2022conversational}, \textit{inter alia}) has significantly advanced the field of natural language generation on tasks including text summarization, machine translation, dialogue generation, etc. Despite the models' increasing capability in fluency, expressiveness and domain generalization, the generated outputs from these models are far from perfect \cite{gehman-etal-2020-realtoxicityprompts,kryscinski-etal-2020-evaluating,fabbri-etal-2021-summeval}. 
As LMs are getting larger and better nowadays, the decoding strategy is another crucial piece in this generation paradigm to release the power of LMs by pulling high probability output sequence from the model.
\begin{figure}[t]
    \centering
    \footnotesize
    \includegraphics[width=0.482\textwidth]{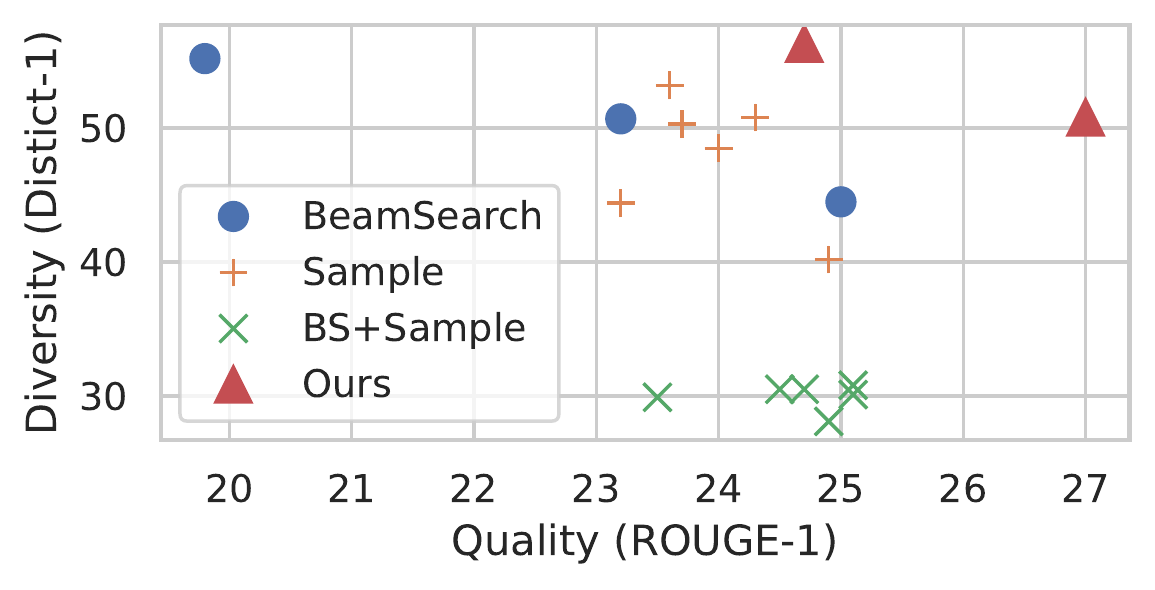}
    \caption{Generated text diversity and quality, measured by Distinctness-1($\uparrow$) and ROUGE-1($\uparrow$), on question generation. The dataset used is QuoRef and the model applied is MixQG. Our approach, best-$k$ search with two configurations, beats popular methods including beam search variations, sampling methods and BS+Sample methods on diversity and quality. See Sec.~\ref{sec:qg} for details.}
    \label{fig:quoref}
\end{figure}
If we form text generation as a search problem, decoding strategies are essentially search algorithms over the space composed by vocabulary $\mathcal{V}$. 
Beam search, a pruning-based heuristic search algorithm, has been the go-to choice for many years. However, the generated sequences are usually repetitive because many diverse and valuable hypotheses are pruned at earlier stage of search \cite{eikema-aziz-2020-map}. 
Sampling-based approaches \cite{fan-etal-2018-hierarchical,Holtzman2020The} can indeed generate more diverse sequences, they are hard to control due to their stochastic nature. Sometimes outputs are duplicate; sometimes a sampling choice with high variance breaks the whole sequence. It also does not provide flexibility to expand a node earlier than the current step, similar to the pruning effect in beam search. 

We are looking for a decoding algorithm with high flexibility and controllability while it could also yield diverse outputs for certain use cases. We find that best-first search (BFS) algorithm is ideal, satisfying these properties. First, as a deterministic algorithm, it doesn't introduce extra variance.
More importantly, since it does not prune hypotheses, it preserves a more diverse set of options and allows simultaneous expansion of hypotheses with different length. 
Despite these intriguing features, we identified some challenges of applying it to text generation after a preliminary study. The search efficiency and search completeness are two major factors hindering the application of BFS. 

Therefore in this work, we propose the \textbf{Best-$k$ Search Algorithm} for diverse and high-quality text generation. 
Our approach re-invents BFS with a few design changes to overcome the issues mentioned before. 
Parallel exploration is designed to explore the top $k$ nodes from the search frontier each time instead of one in BFS. 
It speeds up BFS by around 10x and makes it comparable to beam search in terms of efficiency. 
We also add a temporal decay mechanism to the algorithm to encourage search completions. A simple yet effective stateless scoring function as an alternative to more complicated length-adjusted counterparts is devised, and we show that it works well and helps further in finding diverse texts.

To verify the proposed algorithm, we conduct comprehensive experiments on four tasks, question generation, commonsense generation, text summarization and machine translation. Our results show that the proposed algorithm works well with a wide range of models on six datasets. Our approach yields high-fidelity, diverse and natural outputs while maintaining quality. 
Our contributions are (1) investigation of best-first search for text generation; (2) proposing an efficient, simple, and deterministic decoding algorithm, bestk-$k$ search; (3) comprehensive experiments and strong results on six datasets with ablation study and analysis; (4) The algorithm is free of training or parameter, lightweight, easy-to-use, and compatible with any LLM. It is also orthogonal to many decoding techniques like sampling or rollout. 

\section{Revisiting Best-First Search}
\label{sec:revisit}

In this section, we will introduce the vanilla best-first search in the context of natural language generation as a decoding algorithm, and cover the first Research Question:\textbf{ Is BFS a good algorithm for searching hypotheses in text generation?} 

\paragraph{Setup}
Text generation can be formulated as a sequence generation process given input $\mathbf{x}$ and a probabilistic language model\footnote{Language models (LM) discussed in this paper include unconditional and conditional models, where decoding algorithms could be applied ubiquitously.  } parameterized by $\theta$. 
\begin{align}
    p_{\theta}(\mathbf{y}| \mathbf{x} ; \theta) = \prod_{t=1}^{T} p_{\theta}(y_t | \mathbf{y}_{<t}, \mathbf{x})
\end{align}
Traditionally, maximum \textit{a posteriori} (MAP) decoding strategy is deployed to elicit highest-scoring output sequences $\argmax_{\mathbf{y^{ \ast} }} p_{\theta}( \mathbf{y^{ \ast} }| \mathbf{x})$. Most previous work uses the log-likelihood of the sequence as the proxy for assessing the (partial) sequence quality. However, recent studies found discrepancies between model likelihood and quality assessed by humans \cite{stahlberg-byrne-2019-nmt,Holtzman2020The,eikema-aziz-2020-map,zhang-etal-2021-trading}. 
Various approaches including length normalization \cite{wu2016google}, quality-aware decoding \cite{fernandes-etal-2022-quality}, and regularized decoding \cite{meister-etal-2020-beam} have attempted to modify the objective to mitigate the gap. 
In this work, we adopt $h(\cdot)$ as the scoring function, and $h(\mathbf{y_{1 \cdots t}})$ is the score of a hypothesis $\mathbf{y_{1 \cdots t}}$. 

\paragraph{Graph Notation} 
We frame the derivation of sequences as the expansion of a directed search graph, where BOS is the root node and EOS nodes are the leaf nodes. Any node $n$, except the root node, has exactly one parent node.
The \emph{score} of each node $n$ is defined as the score of the hypothesis starting with BOS and ending with $n$. $h(\cdot)$ abstracts any scoring function.
 Each node $n$ can be represented as a triplet $\langle s, w, t \rangle$ where the score is $s=h(n)$, token $w \in \mathcal{V}$ is the generated token, and $t$ is the time of discovery. A completed sequence is defined as $\hat{y} = (\text{BOS}, \cdots, \text{EOS})$, and $\hat{Y}$ consists of all completed sequences.
 The open-set $\mathcal{O}$ of the graph, also known as the search frontier, is essentially a priority queue\footnote{We use a max-heap for notation simplicity.}.
\begin{algorithm}[t]
\caption{Best-first search}
\label{algo-bfs}
\begin{algorithmic}[1]
\Require Language model abstracted as $p_{\theta}$, search budget, and frontier $\mathcal{O}$.
\Ensure All completed paths $P$ 
\State $\mathcal{O} \leftarrow \{ \langle \infty$, BOS, $-1 \rangle$\}, $T \leftarrow 0$, $t \leftarrow 0$.
 \While{ $T<\text{budget}$ }
 \State $ n \leftarrow \mathcal{O}$\texttt{.pop()}
\For{ $v \in \mathcal{V}$}
\If{ is-complete($n \circ v$)}
\State $P \leftarrow P \cup (n \circ v) $
\State \textbf{continue}
\EndIf
\State child $\leftarrow \langle h(n \circ v), v, t \rangle$
\State  $\mathcal{O} \leftarrow  \mathcal{O} $ $ \cup $ child 
\EndFor
\State    $ T \leftarrow T + 1$
\State $t \leftarrow t + 1$
\EndWhile
\end{algorithmic}
\end{algorithm}

\paragraph{Best-First Search}

Best-first search (BFS) is a greedy search algorithm which explores the graph according to the scoring function $h(\cdot)$. We describe the best-first search algorithm in the context of probabilistic NLG in Algorithm~\ref{algo-bfs}. For each iteration, BFS finds the most promising, expands it, adds newly discovered nodes to $\mathcal{O}$, and repeats until reaching the budget. \emph{is-complete} is the conditional function for termination. $P$ contains completed sequences. $T$ counts the number of explored nodes. 
Although BFS has been applied in many NLP applications \cite{och-etal-2001-efficient,klein-manning-2003-parsing,bostrom2022natural,saha2022summarization}, it is not a popular choice for text generation. Recent work in decoding strategies \cite{meister-etal-2020-best,lu-etal-2022-neurologic,xu-etal-2022-massive} was inspired and motivated by BFS, but none of them directly adopts BFS as the decoding algorithm. 

\begin{table}[t]
\centering
\footnotesize
\narrowcol
\begin{tabular}{@{}r|cccc@{}} 
\toprule
Property & Det. & No Dup. & No Pruning & Diversity\footnote{Diversity is judged by empirical observations, including those from prior work \cite{fan-etal-2018-hierarchical,Holtzman2020The} and this work.}  \\ \midrule
BS       &      \cmark  &  \cmark   &   \xmark     &   \xmark   \\
Sample   &    \xmark    &    \xmark &   \xmark     &   \cmark   \\
BFS      &       \cmark &   \cmark  &   \cmark     &   \cmark    \\ \bottomrule    
\end{tabular}
\caption{Property comparison of search algorithms and approaches. \textit{Det.} stands for deterministic search. \textit{No Dup.} indicates the approach could guarantee no duplication  of output sequences. Duplication and diversity of sampling methods depend on the choice of hyper-parameter.}
\label{tab:bfs-adv}
\end{table}

\subsection{Advantages}
What are the potential advantages of using BFS, compared to beam search and sampling approaches? We enumerate the inherent property of beam search, sampling, and best-first search in Table~\ref{tab:bfs-adv}. BFS has many strengths to satisfy desired properties like diversity, quality, and controllability in text generation. 
\paragraph{Deterministic} BFS is a deterministic search algorithm with lower variance and higher controllability than stochastic sampling methods. This also indicates that BFS is compatible with sampling on top, similar to beam search. 
\paragraph{No duplication} BFS comes with no duplication, so it's guaranteed that the more search budget used, the more unique outputs there will be. Sampling methods with low truncation thresholds suffer from this issue. 

\begin{figure}[t]
    \centering
    \footnotesize
    \includegraphics[width=0.48\textwidth]{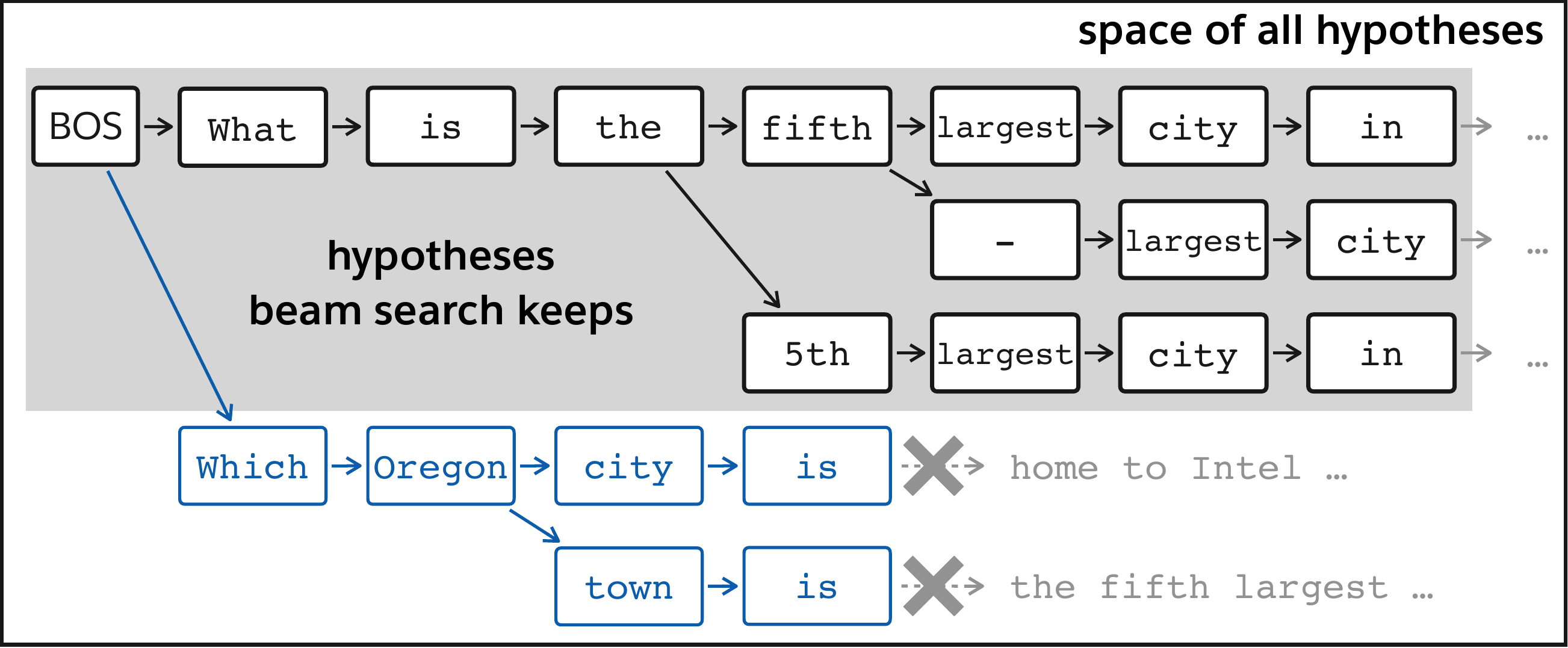}
    \caption{Pruning in beam search removes diverse hypotheses and reduces flexibility of search. This is an example of question generation and the reference is \emph{What city is Intel located in?}. Hypotheses in blue rectangle were discovered but pruned. A greedy completion for \emph{Which Oregon city is} contains information we want, \emph{Intel}. Our approach is able to find and preserve the keyword \emph{Intel}, as shown in Table~\ref{tab:example-quoref}. }
    \label{fig:flex}
\end{figure}

\paragraph{No Pruning} We illustrate the pruning issue in beam search in Figure~\ref{fig:flex}. BS prunes the desired hypotheses. Unlike beam search, BFS never prunes\footnote{In practice, due to the large vocabulary $\mathcal{V}$, we only keep the highest $k$ out of $|\mathcal{V}|$ ranked options for each expansion for efficiency. We posit that the long-tail low probability continuations won't be prioritized by the priority queue and it's fine to discard them anyway.}, and preserves all explored nodes. This also brings great flexibility that the generation could switch between different branches of search. 
\paragraph{Diversity} BFS yields diverse outputs with decent quality. The diversity of generated sequences is based on empirical lens, which will be covered in our experiments.

\begin{table}[t]
\centering
\footnotesize
\narrowcol
\begin{tabular}{@{}r|cccc@{}}
\toprule
Beam Size        & 1      & 2      & 5     & 10    \\ \midrule
Incompletion Rate        & 58.1\% & 23.8\% & 3.9\% & 3.0\% \\
Time (s) & 1.0    & 1.9    & 5.6   & 13.7  \\ \bottomrule
\end{tabular}
\caption{Search incomplete rate and speed for the vanilla BFS. Beam size denotes the equivalent beam size, which is a reflection of the total search budget. Incomplete rate measures how often a search does not reach any completed state (EOS). Time denotes the running time for running the search algorithm per example. }
\label{tab:prelim}
\end{table}

\subsection{Challenges}
\label{sec:challenge}
As we have discussed many strengths BFS enjoys, why has it not been the dominant approach?
We implement a standard best-first search algorithm, as described in Algorithm~\ref{algo-bfs}, and look into how it works on decoding text summaries from \texttt{BART-XSum}\footnote{We describe the technical detail and hardware specification in Appendix~\ref{app:init}. }. We also define a notion of equivalent beam size\footnote{Beam size and equivalent beam size are interchangeable for the rest of the paper for simplicity. We follow \citet{xu-etal-2022-massive} for the definition of equivalent beam size. } to calibrate the search budget for all methods. For beam search, we set a beam size $b$ and a max decoding length $T$, and the total search cost is $C=bT$, which means there will be $C$ times forward passes through LM. BFS can also call the LM for $C$ times and discover $C$ nodes. 

\paragraph{Completedness} 
While beam search iteratively gains depth, best-first search does not. Hence, we investigate how often BFS could (not) reach the search goal, which is at least one EOS token. In Table~\ref{tab:prelim}, we show that the vanilla BFS has a pretty high chance of failure when the search budget is very limited. Even in the case of beam size $b=10$, there is a 3\% of chance that the method won't reach any completed sequence, a sequence ending with EOS or other pre-defined termination tokens. This indicates that \textbf{the vanilla BFS struggles with the completedness}.

\paragraph{Efficiency}
Efficiency is a crucial factor for practical usage. We measure the time consumed for running the search for each example and report it in Table~\ref{tab:prelim}. For reference, beam search with $b=10$ can be completed in 0.7s per example. The vanilla BFS is slow since the step-wise exploration in BFS is not paralleled and batched.


\begin{figure*}[t]
    \centering
    \footnotesize
    \includegraphics[width=\textwidth]{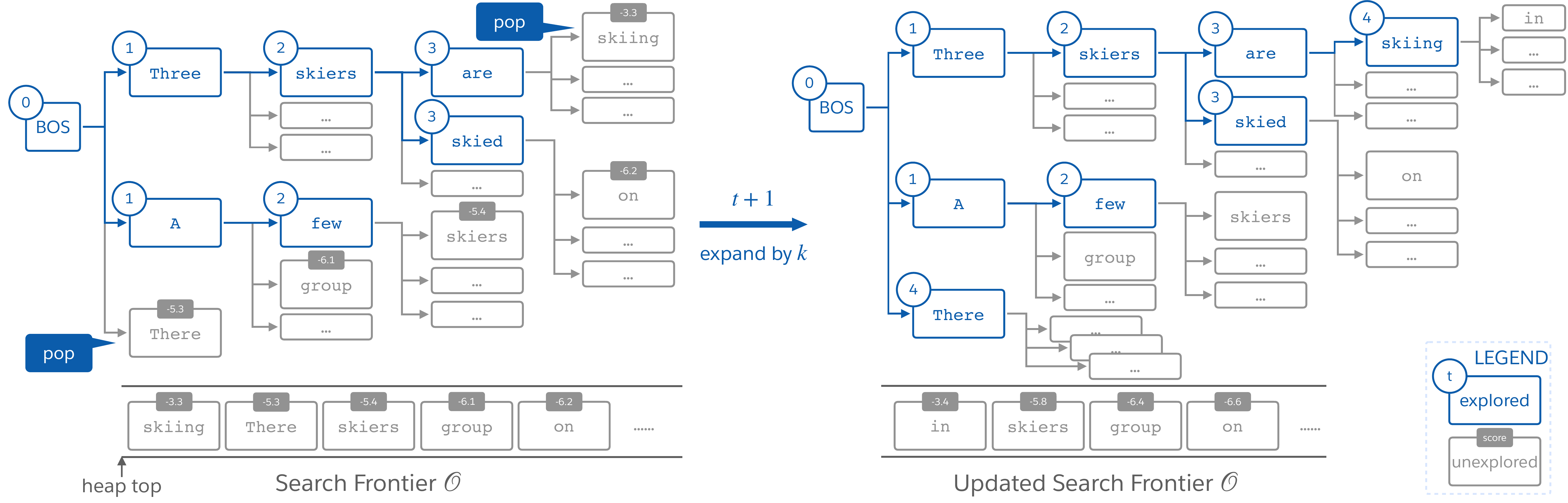}
    \caption{Illustration of best-$k$ search with an example from CommonGen, where the input is ``mountain ski skier". (left) the search graph before expansion; (right) the search graph after expansion with ``skiing'' and ``There'' expanded; (bottom) the search frontier. The upper left number of explored nodes (blue-bordered rectangle) indicates the time stamp of expansion. Grey rectangles are unexplored nodes in the frontier.  For demonstration purposes, we set $k=2$, and only show the top 3 expansions for each node. }
    \label{fig:main}
\end{figure*}
\section{Our Approach: Best-$k$ Search}
In this section, we will introduce Best-$k$ Search, a novel search algorithm inspired by the vanilla best-first search. 
It features a few components: (1) \colorb{parallel exploration}  enables batch-wise exploration in the search graph; (2) \colorg{temporal decay} yields a higher completion rate and fewer dangling nodes; (3) \colorr{heap pruning} improves the time and space efficiency of our approach; (4) a new scoring function which handles various length and promotes the diversity.
We describe the algorithm in Algorithm~\ref{algo:bks} and illustrate it in Figure~\ref{fig:main}. 


\begin{algorithm}[t]

\caption{Best-$k$ Search with \colorb{parallel exploration}, \colorr{heap pruning}, and \colorg{temporal decay}. }
\label{algo:bks}
\begin{algorithmic}[1]

\Require Generation model $\theta$ with vocabulary $\mathcal{V}$, search budget, $\mathcal{O}$ denotes open set (max priority queue). \colorb{group size $k$}. $T$ is the number of explored steps; $t$ is the time stamp.
\Ensure All completed paths $P$.
\State $\mathcal{O} \leftarrow \{ \langle \infty$, BOS, $-1 \rangle$\}, $T\leftarrow0$, $t\leftarrow 0$.
 \While{ $T<\text{budget}$ }
\State $\mathcal{PQ} \leftarrow \emptyset$
 \For{ $ n \in \mathcal{O} $ }
\State $ \mathcal{PQ} \leftarrow \mathcal{PQ} + \langle n\texttt{.score} + \colorg{\texttt{decay} ( n\texttt{.time}, t)  }, n \rangle $
\EndFor
\State \colorb{ $g \leftarrow min(k, \mathcal{PQ}\texttt{.size()})$}
\State  \colorb{$\mathcal{H} \leftarrow  \mathcal{PQ}\texttt{.heappop(g)} $} \Comment{$\mathcal{H}$ is the group of candidates to explore. }
\State $\mathcal{O} \leftarrow \mathcal{O}\setminus \mathcal{H}$

\For{$ \langle score, n  \rangle  \in \mathcal{H}$}

\For{ $v \in \mathcal{V}$}
\If{ is-complete($n \circ v$)}
\State $P \leftarrow P \cup (n \circ v) $
\State \textbf{continue}
\EndIf
\State child $\leftarrow  \langle h( n \circ v), v,  t \rangle $ \Comment{\colorg{Current time $t$ of adding the node to $\mathcal{O}$.} }
\State  $\mathcal{O} \leftarrow  \mathcal{O} \cup $ child 
\EndFor
\EndFor
\State  \colorr{ $\mathcal{O} \leftarrow \mathcal{O}\texttt{.prune()}$}
\State  \colorb{  $T \leftarrow T + g$}
\State    $t \leftarrow t + 1$
\EndWhile
\end{algorithmic}
\end{algorithm}

\subsection{Parallel Exploration}
As suggested in Table~\ref{tab:prelim}, the wall clock running time of BFS is one order of magnitude slower than beam search under similar conditions. Given the same search budget, BFS is supposed to achieve similar time efficiency theoretically. However, multiple step-by-step operations are practically much slower than a batchified operation when GPUs are engaged. 
Hence, we propose a parallel exploration strategy to reduce the exploration time cost by popping $k$ nodes from the priority queue each time and executing them in a batch. 
Current candidates are stored in the frontier $\mathcal{O}$. $\mathcal{PQ}$ is a priority queue after applying any scoring function to nodes in $\mathcal{O}$. 
\begin{equation}
   \mathcal{H} \leftarrow  \mathcal{PQ}\texttt{.heappop(g)} 
\end{equation}
where $g=min(k, \mathcal{PQ}\texttt{.size()})$. 
The strategy serves as an approximation to best-first search as we not only pop the most promising node but also pop the top-$k$ most promising nodes. 
This technique significantly improves the efficiency of best-$k$ search compared to BFS, which will be discussed in Sec.~\ref{sec:eff}. 

\paragraph{Complexity}
With parallel exploration, the number of batches fed to GPU is $\frac{b}{k}T$. For beam search, it is $T$; for BFS, it is $bT$. 
We set the cost of beam search for each batch as a unit cost, which accounts for calling LM and other operations. 
Let the extra overhead cost of best-$k$ search for each batch as $\epsilon$, which accounts for padding, heap pruning, node handling, etc. 
The time complexity of best-$k$ search is $\mathcal{O}_{best-k} = \frac{b}{k}T(1+\epsilon)$ where for beam search it is $\mathcal{O}_{BS} = T$.
What is worth noticing is the value of $k$ is supposed to be bounded by beam size $b$; otherwise, the actual maximum decoding depth $\frac{b}{k}T$ is smaller than $T$.
In our experiments, we set $k=5$ or $k=10$ when the beam size $b=10$.

\subsection{Temporal Decay}
Completion, measured by the number of outputs from the algorithm, has been another key challenge for BFS. In Table~\ref{tab:prelim}, increasing the search budget helps improve the completion rate but there is still a non-trivial portion of samples that fails. We propose a technique to fulfill the completition goal during the search process. 
For each node added to the search frontier $\mathcal{O}$, we keep the time stamp $t$. When we pop nodes, we modify the score of each node by adding an auxiliary score rewarding recently discovered nodes. The idea is to increase the score of recently discovered nodes so the algorithm prefers to continue them.
The decay function needs to be monotonic. Hence, we define the decay function as a power function:
\begin{equation}
    \texttt{decay}(n\texttt{.time} , t) = -\kappa ( t - n\texttt{.time} )^{\beta}
\end{equation}
where $\kappa>0$ controls the weight of the term and $\beta>0$ controls the slope. $t$ is the current time step and $n\texttt{.time}$ is a past time step, so $t - n\texttt{.time} > 0$. The older the node, the smaller the value of $\texttt{decay}(n\texttt{.time} , t)$. A more recent node will receive a higher incentive from the decay function, so it's more likely to be popped and expanded. 
For example, a node discovered at $t=1$ receives $\texttt{decay}(1, 5)=-4$ and a node discovered at $t=4$ receives $\texttt{decay}(4, 5)=-1$, if we set $\kappa=\beta=1$.
In our experiment, we set $\beta=0.5$ and explore different values of $\kappa$.
We leave other forms of the decay function, i.e. logarithm, as future work. We discuss some design choices for  competition in Appendix~\ref{app:len}.

\subsection{Heap Pruning}
The size of the heap grows fast during exploration. For most of the time, however, our approach only utilizes top-ranked hypotheses. 
The temporal decay function is monotonic, so for any node in the search frontier, the final score is always decreasing as the time moves forward. 
The usage of the temporal decay could affect the ranking, but we posit that if the margin of model score between a candidate node and the $k$-th highest node from the heap is larger than $\epsilon$, it is unlikely that it will be used in future. 
The choice of the margin $\epsilon$ depends on factors including the intensity of temporal decay, remaining search budget, model calibration, and resource limitations.
In practice, we set a sufficiently large maximum heap size to 500 to avoid tuning $\epsilon$ on different datasets.

The expansion of each node could lead to $|\mathcal{V}|$ extension nodes, where $|\mathcal{V}|$ is the size of the vocabulary. As the conditional probability $p_{\theta}(y_t | \mathbf{y}_{<t}, \mathbf{x})$ is usually long-tailed, it is fine to discard those low-scoring nodes for space and time efficiency. We set a threshold $\gamma=0.05$ to filter out generations with probability lower than it. The smaller $\gamma$ is, the more nodes we need to instantiate and manage.



\subsection{Model Score}
\label{sec:modelscore}
The depth of a BFS search graph is not aligned while the that of beam search remains the same during the search. As the scoring function plays a crucial role in finding ideal sequences $\hat{Y}$, we investigate whether existing scoring functions are still compatible with the best-$k$ search algorithm. 
Here are a few common ways to define the scoring function $h$ regarding the length $l$ of the (partial) sequence: 
\begin{enumerate}
    \item  original: $h(\mathbf{y}) = \sum_{t=0}^{l} \log p_{\theta}(y_t | \mathbf{y}_{<t}, \mathbf{x} )$. This is the original way of defining the score of a sequence with its sequence log-likelihood. 
    \item  length-adjusted scoring function:  $h(\mathbf{y}) =\frac{1}{|\mathbf{y}|^{\alpha}} \sum_{t=0}^{l} \log p_{\theta}(y_t | \mathbf{y}_{<t}, \mathbf{x} )   $. The tunable hyper-parameter $\alpha$ controls the preference of length  \cite{meister-etal-2020-beam}. 
\end{enumerate}
The hypotheses in BFS have different length so it's tricky to pick a good hyper-parameter for length-adjusted functions across samples and datasets. In this work, we also propose a memoryless scoring function:
\begin{equation}
    h(\mathbf{y}) = \log p_{\theta}(y_t | \mathbf{y}_{<t}, \mathbf{x} )
\end{equation}
It approximates the score of the whole hypothesis $\mathbf{y}$ with the probability of the last node. 
It satisfies the Markov property that only the last state's probability is considered for the next continuation. 
When we use this scoring function together with best-$k$ search, we term the approach as \bfslast{}.
We conduct ablation studies to understand different scoring functions in Sec.~\ref{sec:discuss-score}. 
We found that the length-biased scoring function typically works the best while the memoryless function generates more diverse outputs with slightly lower quality. 

\section{Evaluation}

\subsection{Tasks, Models \& Datasets}

We investigate four conditional text generation tasks, ranging from more precision-oriented tasks like machine translation to more open-ended tasks like commonsense generation and question generation. 
MT is a use case where diverse outputs are not always required, so we devise our algorithm followed by reranking to see how much we can benefit from many diverse and high-quality outputs. We will present the result of machine translation in Section~\ref{sec:mt} 

\paragraph{Question Generation}

Question generation aims to generate questions given the context, and the generated questions further enable applications like data augmentation, question decomposition, and fact verification \cite{fabbri-etal-2022-qafacteval}. We adopt a state-of-the-art question generation model, MixQG \cite{murakhovska-etal-2022-mixqg}, as the testbed to verify whether our approach could elicit more diverse, larger number and high-quality questions compared to baseline approaches. We use the variant \texttt{mixqg-large} in this paper. 
For datasets, we select a range of seen and unseen QA datasets, including SQuAD \cite{rajpurkar-etal-2016-squad}, DROP \cite{dua-etal-2019-drop}, and QuoRef \cite{dasigi-etal-2019-quoref}. Following \citet{murakhovska-etal-2022-mixqg}, we prepend the answer span to the document as the model input. We set the maximum decoding length to 25 BPEs for SQuAD and QuoRef, and 20 for DROP.

\paragraph{Commonsense Generation} CommonGen is a dataset for generative commonsense reasoning \cite{lin-etal-2020-commongen}. The input is a few keywords and the target is a sentence satisfying commonsense and covering these keywords. We adopt a popular T5-based model\footnote{The model is available at \url{https://huggingface.co/mrm8488/t5-base-finetuned-common_gen}.} fine-tuned on the training set of CommonGen. Since CommonGen has multiple references for each input, we utilize multiple references for each example by evaluating outputs against them. The maximum decoding length is set to 20. 


\paragraph{Text Summarization}
We use XSum \cite{narayan-etal-2018-dont} as the dataset for abstractive text summarization. The model we use for this task is the BART\footnote{\url{https://huggingface.co/facebook/bart-large-xsum}} model \cite{lewis-etal-2020-bart} fine-tuned on XSum. The maximum decoding length is 30.

\subsection{Baselines}

\textbf{Beam earch} (\bs) is the long-standing choice for decoding sequences for decades \cite{reddy1977speech} and \textbf{diverse beam search} is a diversity-promoting variant of beam search \cite{vijayakumar2016diverse}. We experiment with different numbers of beam groups for diverse beam search: 5 for \dbs{} and 10 for \dbsplus.
\textbf{Sample} is represented by two widely-adopted strong stochastic sampling methods, nucleus sampling (\textsc{Ncls}) \cite{Holtzman2020The} and typical sampling (\textsc{Typ}) \cite{meister-typical}. \textbf{Beam sample} includes a collection of beam search multinomial sampling methods. We experiment with the integration of beam search with typical sampling and nucleus sampling, denoted as \textsc{BNcls} and \textsc{BTyp} respectively. Implementation of baseline approaches is available at \href{https://huggingface.co/docs/transformers/main/en/main_classes/text_generation#transformers.GenerationMixin.generate}{Transformers/GenerationMixin/generate}.


\paragraph{Ours} We use two typical configurations to represent our approach: \bfslast where the scoring function is memoryless, and \bfsmean where $\alpha=1$. In \bfsmean, the score of the sequence is the average log-likelihood of individual time steps. We experiment with $k=\{5, 10\}$ and the weight of temporal decay in $\{0.0, 0.01, 0.05, 0.1, 0.2\}$, and report the configuration with the best combination of diversity (\ad) and quality (\ar). 
\begin{table*}[t]
\centering
\footnotesize
\begin{tabular}{@{}r|cr|ccc|ccc|c|ccccc@{}}
\toprule
             & \multicolumn{2}{c|}{Stat} & \multicolumn{3}{c|}{Diversity ($\uparrow$)}           & \multicolumn{3}{c|}{ Oracle ($\uparrow$) } & Natural ($\uparrow$) & \multicolumn{5}{c}{Quality ($\uparrow$)}            \\
Method       & \textsc{S}      & |S|    & \done & \dtwo & \dthree & \textsc{R1}         & \textsc{R2}                 & \textsc{RL}                 & \mv   & \textsc{R1}         & \textsc{R2}                 & \textsc{RL}     & \mtr & \grm \\ \midrule
BS           & 10      & 10            & 44.8        & 48.7        & 46.9        & 32.6        & 12.9       &  \finecell 30.1       & 59.5    & 25.9 & \goodcell 9.2 & \goodcell 23.7 & \goodcell 20.9   & 88.9      \\
DBS          & 10       &   9             &  52.3        & 52.2        & 47.6        & 30.1        & 9.5        & 26.4       & 41.5    & 24.2 & 7.3 & 21.3 & 18.7   & \badcell  85.2      \\
DBS+         & 10       & 9   &  \goodcell 55.8        & 53.1  & 45.8 & 26.1     & \badcell  6.8        & 23.4       & \badcell  13.7    &  \badcell 20.3 &  \badcell 4.5 & \badcell  17.8 & \badcell  14.9   & 85.7      \\ \midrule
\btyp{0.2} & 10 & 1 & 29.9 & 27.8 & 24.3 & \badcell  24.2 & 7.0  & \badcell  22.0 & 53.5 & 23.5 & 6.7 & 21.4 & 18.3 & 90.8 \\
\btyp{0.5} & 10 & 2 & 30.5 & 28.4 & 24.7 & 25.0 & 7.5  & 22.7 & 48.1 & 24.7 & 7.2 & 22.4 & 19.2 & \finecell 92.5 \\
\btyp{0.95} & 10 & 2 & 30.9 & 28.9 & 25.4 & 26.9 &	8.6 &	24.8  & 61.4 & 25.0	& 7.6 &	23.1 & 19.5 & 92.3 \\
\btopp{0.5} & 10 & 1 & \badcell 28.1 & \badcell  25.0 & \badcell  21.1 & 24.9 & 7.0  & 22.7 & 51.2 & 24.9 & 7.0 & 22.8 & 18.2 & 92.3 \\
\btopp{0.8} & 10 & 2 & 30.1 & 28.0 & 24.4 & 25.6 & 7.9  & 23.7 & 49.9 & 25.1 & 7.3 & 23.1 & 19.1 & \finecell 92.5 \\
\btopp{0.9} & 10 & 2 & 30.8 & 28.7 & 25.2 & 26.0 & 8.2  & 24.0 & 58.6 & 25.1 & 7.5 & 23.0 & 19.3 & 91.2 \\ \midrule
\typ{0.2}   & 10 & 5 & 44.4 & 46.2 & 42.4 & 26.5 & 7.3  & 23.9 & 50.9 & 23.2 & 6.3 & 21.1 & 18.1 & 88.3 \\
\typ{0.5}    & 10 & 7 & 48.5 & 52.0 & 47.9 & 30.1 & 10.6 & 27.3 & \finecell 71.0 & 24.0 & 6.5 & 21.9 & 18.0 & 91.8 \\
\typ{0.95}    & 10 & 9 & 54.3 &	59.4	&55.7	&31.2&	11.2&	28.5 & 84.3 & 22.1 & 6.1 & 20.1 & 17.1 & 89.5 \\
\topp{0.5}    & 10 & 5 & 40.2 & 41.4 & 37.7 & 29.2 & 9.9  & 26.3 & 58.3 & 24.9 & 7.3 & 22.9 & 18.6 & \goodcell  93.9 \\
\topp{0.8}    & 10 & 8 & 50.8 & 55.1 & 51.3 & 30.5 & 10.2 & 27.3 & 47.7 & 24.3 & 6.2 & 21.6 & 18.2 & 91.1 \\
\topp{0.9}    & 10 & 9 & 53.2 & \finecell 58.2 & 53.7 & 31.2 & 11.5 & 28.7 & 46.0 & 23.6 & 6.9 & 21.4 & 18.0 & 90.9  \\ \midrule
MixQG     & -       & -      &     -     &      -    &     -     &      -    &      -    &    -     &    -  & 24.9 & 8.0 & 22.3 & -  & -      \\
\bfsmean    & 20       & 20            & 50.8 &	56.1 & \finecell	54.0       &  \goodcell  33.9	& \goodcell 14.2 & \goodcell	31.0       & \goodcell 83.0    & \goodcell 27.0	& \finecell 8.9 & \goodcell	24.5 & \goodcell 21.3  & 86.8   \\
\bfslast & 19  & 19   & \finecell 53.4   & \goodcell 59.4        & \goodcell 55.9        & \finecell  32.7        & \finecell  13.4       & \finecell 30.1       & 69.4    & \finecell  26.0 & 8.4 & 23.2 & 19.7   & 91.7     
\\ \bottomrule
\end{tabular}
\caption{Experiments result on QuoRef question generation. S and |S| stand for the number of sentences and the unique number of sentences. \textsc{D}-1, -2, and -3 stand for unigram, bigram and trigram distinctness. \textsc{Mv} is the \textsc{Mauve} score measuring the naturalness of the generated outputs. \textsc{Mtr} is METEOR score. \textsc{Grm} measures the grammaticality. We highlight the \colorbox{RoyalBlue!25}{best}, \colorbox{RoyalBlue!10}{second best}, and the \colorbox{BrickRed!15}{worst} for each column.  A visualized comparison with \done{} and \textsc{R1} is presented in Figure~\ref{fig:quoref}.}
\label{tab:quoref}
\end{table*}

\subsection{Metrics}
We measure the generated outputs from multiple aspects including text quality, relevance, diversity, and naturalness.

\paragraph{Statistics}
We report the number of completed strings and the number of unique completed strings as S and |S|.

\paragraph{Diversity} 
Following \citet{li-etal-2016-diversity,yang-klein-2021-fudge}, we report the distinctness of completions, measured as the number of unique $n$-grams divided by the number of words, denoted as \done{}, \dtwo{} and \dthree.

\paragraph{Text Quality}
We adopted two relevance based metrics, ROUGE (\textsc{R1}, \textsc{R2}, \textsc{RL}) \cite{lin-2004-rouge} and METEOR (\mtr) \cite{banerjee-lavie-2005-meteor}, for assessing the surface similarity between the generated strings and the reference. 

\paragraph{Naturalness}
We measure the naturalness of the generated sequences with MAUVE \cite{pillutla2021mauve}, a metric for open-ended text generation.

\begin{table}[t]
\centering
\footnotesize
\narrowcol
\begin{tabular}{@{}r|rcccccc@{}}
\toprule
 & |S| & \ad  & \ao & \ar & \mv & \grm & \mtr  \\ \midrule
\bs          & 10        & 23.0 & 34.2 & 25.1 & 9.6           & 88.1         & 29.8           \\
\dbs          & 9         & 26.6 & 32.8 & 23.1 & 13.0          & 82.3         & 27.9           \\
\dbsplus          & 9         & 30.9 & 32.6 & \badcell 19.8 & \badcell 9.0           & \badcell 80.6         & \badcell 23.1           \\ \midrule
\btyp{0.2}          & 1         & 9.7  & \badcell 25.6 & 25.1 & 15.6          & 88.4         & 29.7           \\
\btyp{0.5} & 1 &     10.3 &	25.8 &	25.0	 &  \goodcell 36.2 &	\goodcell 93.4 &	30.1 \\
\btyp{0.95}          & 2  &     11.0 & 	28.0 &  \goodcell	26.6 & 	11.9 & 	89.2 &  \goodcell	31.1        \\
\btopp{0.5}          & 1         & \badcell 9.2  & 26.7 & 26.4 & 13.0          & \finecell 90.2         & 30.7           \\
\btopp{0.8}          & 2         & 10.5 & 27.5 & \finecell 26.5 & 9.3           & 89.6         & \finecell 30.9           \\
\btopp{0.9}          & 2         & 10.8 & 27.9 & \finecell 26.5 & 10.0          & 89.2         & \finecell 30.9           \\ \midrule
\typ{0.2}          & 6         & 22.6 & 29.1 & 23.7 & 17.0          & 86.6         & 28.1           \\
\typ{0.5}          & 8         & 28.0 & 34.8 & 24.9 & 13.3          & 88.7         & 29.4           \\
\typ{0.95} & 10&  \finecell 36.2&	34.9&	23.3&	18.8	&84.0&	27.8 \\
\topp{0.5}          & 5         & 18.7 & 32.1 & \finecell 26.5 & 13.8          & 89.8         & 30.8           \\
\topp{0.8}          & 9         & 30.2 & 35.4 & 24.8 & 16.5          & 86.6         & 29.3           \\
\topp{0.9}          & 9         & 33.8 & 35.6 & 24.0 & 16.2          & 86.2         & 28.6           \\ \midrule
\bfsmean          & 29        & 30.3 & \finecell 35.8 & 25.5 & 16.5          & 88.6         & 30.5           \\
\bfslast & 24	& \goodcell 36.5	& \goodcell 36.1	& 21.7	& \goodcell 22.6 & 	86.4 &	25.8 \\
\bottomrule
\end{tabular}
\caption{Results of question generation on DROP. \ad{} is the average of \done{}, \dtwo{} and \dthree{}. \ao{} and \ar{} are the average of Oracle ROUGE and ROUGE. }
\label{tab:drop}
\end{table}

\subsection{Question Generation}
\label{sec:qg}
For QuoRef and DROP, we present the experiment results in Table~\ref{tab:quoref} and \ref{tab:drop} respectively. 
Due to the space limit, we present the results of SQuAD in Table~\ref{tab:squad} in Appendix~\ref{app:squad} and the condensed result for SQuAD and DROP.  
On QuoRef question generation, our approach has overall the best diversity, oracle ROUGE, naturalness, and text quality, compared to all baselines. Diversity-oriented methods like \dbs{} and \dbsplus{} do not provide decent text when pursuing diverse texts. 
Compared to beam search, \bfslast and \bfsmean achieve similar ROUGE score but double the number of outputs. 
Our methods also achieve significantly higher \textsc{Mauve} score than peer methods. 
To visualize the \emph{trade-off} in quality and diversity, we also visualize these two metrics in Figure~\ref{fig:quoref}, which shows our approach significantly surpasses all baseline methods on both diversity and text quality, measured by \done{} and \textsc{R1}. There is a typical trade-off curve for diversity and quality by controlling hyper-parameters ($p$ value for nucleus sampling, group size for diverse beam search, etc.), but our approaches go beyond the established curve by a significant margin. 

For DROP, our approaches outperform baseline models in either diversity or quality. For example, on DROP, \btopp{0.9} achieves $\overline{\textsc{D}}=10.8$ and $\overline{\textsc{R}}=26.5$ while \bfsmean comes with $\overline{\textsc{D}}=30.3$ and $\overline{\textsc{R}}=25.5$. We posit that in many real-world applications, a gap of 0.5 average ROUGE score will not change the whole story, but 10x more output (2 vs. 29) and substantial gain of diversity (10.8 vs. 30.3) could unlock tons of applications and choices.

\begin{table}[t]
\centering
\footnotesize
\narrowcol
\begin{tabular}{@{}r|rcccccc@{}}
\toprule
        & |S| & \ad  & \ao & \ar & \mv & \grm & \mtr \\ \midrule
\bs                         & 10       & 40.6 & 42.1 & 40.3 & 23.4      & 88.3     & 42.7       \\
\dbs                        & 10       & 48.2 & \finecell 42.6 & 37.9 & 21.6      & 79.2     & 37.3       \\
\dbsplus                    & 10       & 54.1 & 42.4 & 36.4 & 15.9      & 77.5     & \badcell 35.8       \\ \midrule
\btyp{0.2}                   & 2        & 27.4 & \badcell 36.3 & 38.0 & 27.0      & 83.8     & 40.0       \\
\btyp{0.5}                   & 2        & 26.7 & 37.7 & 40.4 & 17.1      & 88.9     & 43.0       \\
\btyp{0.95}                   & 2        & 27.9 & 38.4 & 40.7 & 14.6      & 89.2     & 43.3       \\
\btopp{0.5}                  & 1        & \badcell 24.3 & 37.1 & 40.5 & \badcell 11.9      & 87.9     & 43.1       \\
\btopp{0.8}                  & 2        & 27.0 & 38.5 & \goodcell 41.0 & 16.9      & \finecell 89.5     & \finecell 43.5       \\
\btopp{0.9}                  & 2        & 27.4 & 38.5 & \finecell 40.9 & 15.4      & \goodcell 89.6     & \goodcell 43.6       \\ \midrule
\typ{0.2}                    & 9        & 55.4 & 40.5 & 34.9 & 37.9      & 79.3     & 37.8       \\
\typ{0.5}                    & 10       & 55.0 & 42.4 & 37.1 & 37.8      & 82.2     & 39.3       \\
\typ{0.95}                    & 10       & \goodcell 61.0 & 39.5 & \badcell 33.7 & \goodcell 41.7      & \badcell 74.9     & 36.1       \\
\topp{0.5}                   & 8        & 44.6 & 41.1 & 39.2 & 24.6      & 86.2     & 41.5       \\
\topp{0.8}                   & 10       & 55.7 & 41.7 & 36.5 & 31.7      & 82.0     & 38.6       \\
\topp{0.9}                   & 10       & \finecell 59.7 & 41.2 & 35.6 & \finecell 41.5      & 79.4     & 38.0       \\ \midrule
\bfsmean     & 27       & 45.7 & 41.4 & 38.2 & 19.2      & 83.8     & 41.0       \\
\bfslast & 22       & 51.4 & \goodcell 43.3 & 37.6 & 34.6      & 84.7     & 39.3      \\\bottomrule
\end{tabular}
\caption{Results on commonsense generation.}
\label{tab:cg}
\end{table}

\subsection{Commonsense Generation}
We present the experimental result of commonsense generation in Table~\ref{tab:cg}. Sampling based approaches are overall good at diversity but the quality of generated text is lower than other methods. For example, \typ{0.95} has the best average distinctness score and the worst ROUGE score at the same time. We observe a trade-off between quality and diversity on this dataset, and our approach yields outputs with a great balance of diversity and quality. Our approach \bfslast{} also has the highest oracle ROUGE score, which indicates high search quality over human annotations.  

\begin{table}[t]
\centering
\footnotesize
\begin{tabular}{ p{0.93\linewidth} } \toprule
\multicolumn{1}{c}{\textsc{GBS} / \textsc{DBA} / \textsc{NeuroLogic$^{\star}$}}                                               \\ \midrule
G: A dog is run over by a ball and mouth agape.                          \\
D: A dog is run over by a ball and bites his mouth.                      \\
N: A dog running with a ball in its mouth.                               \\ \midrule
\multicolumn{1}{c}{\topp{0.8}}                                               \\\midrule
A dog running around with a ball in his mouth.                       \\
The dog is running with a ball in his mouth.                         \\
The dog runs away with the ball out of the mouth.                    \\
A dog running on its mouth with a ball                               \\
A dog with a ball running around his mouth.                          \\ \midrule
\multicolumn{1}{c}{\typ{0.5}}                                               \\\midrule
A dog with a ball in its mouth running around the pond.              \\
A dog runs to the door, eating a ball, and another dog in the mouth. \\
A dog running away with a ball in its mouth.                         \\
A dog running with a ball in his mouth.                              \\
A dog is running around its mouth catching a ball.                   \\ \midrule
\multicolumn{1}{c}{Ours}                                              \\\midrule
A dog is running around with a ball in its mouth.                     \\
a dog running around with a ball in its mouth                         \\
The dogs are running around with balls in their mouths.               \\
Two dogs running around in the same room with a ball in their mouths. \\
Two dogs running with balls in their mouths.                         \\ \bottomrule
\end{tabular}
\caption{An example from CommonGen where the input is ``ball dog mouth run''. We first present the outputs on \textsc{GBS} \cite{hokamp-liu-2017-lexically}, \textsc{DBA} \cite{post-vilar-2018-fast}, and \textsc{NeuroLogic$^{\star}$}, provided in \citet{lu-etal-2022-neurologic}. Then we show five sample outputs from \topp{0.8}, \typ{0.5} and \bfslast{}, respectively.}
\label{tab:example-cg}
\end{table}

\begin{table}[t]
\centering
\narrowcol
\footnotesize
\begin{tabular}{@{}r|rcccccc@{}}
\toprule
 & |S| & \ad  & \ao & \ar & \mv & \grm & \mtr \\ \midrule
\bs           & 8       & 16.3 & 36.6  &  \finecell  31.2 & 98.0  & \goodcell 96.4         &  36.9           \\
\dbs          & 8       & 20.5 & 36.3  & 28.9 & 64.6  & 95.2         & 32.3           \\
\dbsplus         & 7       & 21.5 & 35.6  & 27.8 & \badcell  22.3  & \badcell  92.0         & \badcell 29.8           \\ \midrule
\btyp{0.2}     &  2       & \badcell  12.3 & \badcell  29.5  & 27.6 & \goodcell 98.8  & 96.0         & 34.2           \\
\btyp{0.5}      & 3       & 13.4 & 33.0  & 30.4 & 98.2  & 96.3         & 36.5           \\
\btyp{0.95}      & 3       & 13.3 & 33.7  & 30.9 & 98.5  & \goodcell 96.4         & 37.0           \\
\btopp{0.8}       & 3       & 13.2 & 33.5  & 30.8 & 98.5  & 96.3         & \goodcell 37.1           \\
\btopp{0.9}       & 3       & 14.0 & 34.1  & 31.0 & 98.5  & \goodcell 96.4         & \goodcell 37.1           \\ \midrule
\typ{0.2}  & 7       & 30.9 & 34.2  & \badcell  26.7 & 97.8  & 94.7         & 31.3           \\
\typ{0.5}  & 8       & 34.7 & \finecell 38.8  & 28.8 & 97.9  & 95.1         & 32.7           \\
\typ{0.95}  & 8       & \finecell 35.7 & 38.5  & 28.1 & 98.4  & 95.1         & 32.3           \\
\topp{0.8}  & 8       & 35.3 & \finecell 38.8  & 28.7 & 98.1  & 95.1         & 32.9           \\
\topp{0.9}  & 8       & \goodcell 37.2 & 37.7  & 27.3 & 98.5  & 94.4         & 31.4           \\ \midrule
\bfsmean     &  22      & 21.4 & \goodcell 39.0  & \goodcell 31.9 & \goodcell 99.5  & 95.3         & 35.9           \\
\bfslast &  17     & 24.3 & 37.5  & 28.9 & 98.5  & 95.7         & 33.3           \\ \bottomrule
\end{tabular}
\caption{Results on XSum with BART finetuned on XSum.}
\label{tab:xsum}
\end{table}

\subsection{Text Summarization}
In text summarization, although only one reference is provided, there are many different ways, styles and aspects to summarize an article. As reranking text summarization system outputs has gained increasing interest, the fruitfulness and diversity of generated summaries are valuable attributes to look at.
We present the result of text summarization in Table~\ref{tab:xsum}. Our approach remains competitive in quality, diversity, and naturalness. Our approach achieves an average ROUGE of 31.9 and MAUVE of 99.5, higher than any other methods. \ad{} of our approach is lower than sampling due to the longer sequence lengths and more dangling nodes.

\section{Analysis}

\begin{table*}[t]
\centering
\footnotesize
\narrowcol
\begin{tabular}{@{}p{0.29\linewidth} p{0.34\linewidth} p{0.34\linewidth}@{}} \toprule
\multicolumn{1}{c}{Sampling} &
  \multicolumn{1}{c}{\bfslast{}} &
  \multicolumn{1}{c}{\bfsmean{}} \\ \midrule
\begin{tabular}[c]{@{}p{\linewidth}@{}} \centerline{\textbf{\topp{0.8}} }  What is the fifth largest city in OR?\\ What is the fifth largest city in OR?\\ What is the fifth largest city in OR?\\ What is the fifth largest city in OR?\\ What is the fifth largest city in OR?\\ What is the fifth-largest city in OR?\\ What is the fifth-largest city in OR?\\ What is the fifth-largest city in the State of OR?\\ What is the fifth-largest city in the State of OR?\\ Which city in OR is the county seat of Washington County?\\ \centerline{\textbf{\typ{0.5}}}   What is the fifth largest city in OR?\\ What is the fifth largest city in OR?\\ What is the fifth largest city in OR?\\ What is the fifth largest city in OR?\\ What is the fifth largest city in OR?\\ What is the fifth-largest city in OR?\\ What is the fifth-largest city in OR?\\ What is the fifth-largest city in OR?\\ Which city in OR is the county seat of Washington County?\\ Which city is the county seat of Washington County?\end{tabular} &
  \begin{tabular}[c]{@{}p{\linewidth}@{}}What city is the fifth largest?\\ What city is the fifth-largest city in the State?\\ What is the 5th largest city in OR?\\ What is the fifth largest city in OR?\\ What is the fifth largest city in the State of OR?\\ What is the fifth largest city in the State?\\ What is the fifth-largest city in the State?\\ Which city is the fifth largest city in OR?\\ Which city is the fifth largest city?\\ Which is the fifth largest city?\\ Which OR city is the fifth largest in the state?\\ Which OR city is the fifth largest?\\ Which OR town is home to \colorb{Intel}?\\ Which OR town is home to the tech company \colorb{Intel}?\\ Which OR town is known as the Silicon Forest?\\ Which OR town is the fifth largest city in the state?\\ Which OR town is the fifth largest city?\\ Which OR town is the fifth largest in size?\\ Which OR town is the fifth largest in the state?\\ Which OR town is the fifth largest?\end{tabular} &
  \begin{tabular}[c]{@{}p{\linewidth}@{}}What city in OR is the fifth largest in OR?\\ What city is the fifth largest city in OR?\\ What city is the fifth largest city in the State?\\ What city is the fifth largest in OR?\\ What city is the fifth largest in the state?\\ What city is the fifth largest?\\ What city is the fifth-largest in the State?\\ What is the fifth largest city in OR?\\ What is the fifth largest city in the State?\\ Which city in OR has the largest population?\\ Which city in OR hosts \colorb{Intel}?\\ Which city in OR is known as the Silicon Forest?\\ Which city in OR is the fifth largest in OR?\\ Which city in OR is the fifth largest in the state?\\ Which city is the fifth largest city in OR?\\ Which city is the fifth largest city?\\ Which city is the fifth largest in the state?\\ Which city is the fifth largest?\\ Which OR city is the county seat of Washington County?\\ Which OR city is the fifth largest in size?\\ Which OR city is the fifth largest?\end{tabular} \\ \bottomrule
\end{tabular}%

\begin{tabular}{@{}p{\linewidth}@{}}
Input (Ans || Context): Hillsboro || Hillsboro is the fifth-largest city in the State of Oregon and is the county seat of Washington County. Lying in the Tualatin Valley on the west side of the Portland metropolitan area, the city hosts many high-technology companies, such as \colorb{Intel}, that comprise what has become known as the Silicon Forest. At the 2010 Census, the city's population was 91,611.For thousands of years before the arrival of European-American settlers, the Atfalati tribe of the Kalapuya lived in ... \\
Reference Question: What city is \colorb{Intel} located in?
\end{tabular}%
\caption{Example on QuoRef question generation. We compare the output from sampling methods and our approaches. The input and the reference question are provided at the bottom. The duplication of the sampling approach is high while our model generates a more diverse set of questions. Some outputs from our approach cover the entity \colorb{Intel} mentioned in the reference question. We manually replace all the occurrences of \textit{Oregon} with \textit{OR} due to the layout limit.}
\label{tab:example-quoref}
\end{table*}

\subsection{Example}
We show one example output of CommonGen in Table~\ref{tab:example-cg}. We list outputs provided by \citet{lu-etal-2022-neurologic}\footnote{The model we use is different from the ones in \citet{lu-etal-2022-neurologic}, so their outputs are only for reference.} and the outputs from our experiments. The second output of \typ{0.5}, saying \emph{A dog ... eating a ball, and another dog in the mouth}, makes no sense. The outputs from our model are more diverse since multiple types of subjects exists, including \emph{a dog}, \emph{the dogs}, and \emph{two dogs}. 

We also present one example output from QuoRef question generation. In this example, we can observe the duplication issue rooted in sampling based methods. Most of the generated questions from sampling are duplicate, covering the easiest question to ask. However, our approaches yield diverse and high-quality questions, covering broader spectrum of facts and knowledge like \emph{Intel}, \emph{Silicon Forest}, \emph{country seat of Washington County}. 

\begin{table}[t]
\centering
\footnotesize
\begin{tabular}{@{}r|cccc@{}} \toprule
     & \multicolumn{2}{c}{Best-$k$ Search} & \bs & \btyp{0.5}  \\
     & $k=5$         & $k=10$       &   $b=10$  & $b=10$           \\ \midrule
Time & 1.8s        & 1.2s       & 0.7s          & 1.4s   \\
|S|  & 18.2        & 12.8       & 8.3           &  3.0    \\ \bottomrule
\end{tabular}
\caption{Efficiency comparison of our approach and beam search. Time shows the decoding time used for each example.}
\label{tab:speed}
\end{table}

\subsection{Efficiency}
\label{sec:eff}
We test the wall-clock running time of our algorithms and the standard beam search. We follow the same configuration in Sec.~\ref{sec:revisit}. The result is presented in Table~\ref{tab:speed}. Although our approach is still slower than beam search, due to all the overhead cost including padding sequences, scoring hypotheses and heap management, the speed is reasonable for many applications. The heap size could be shrunk and the heap management could be optimized for even better efficiency. 


\begin{table}[t]
\centering
\small
\begin{tabular}{@{}rcccc@{}}
\toprule
        &  $\alpha=0$ & $\alpha=0.5$ & $\alpha=1.0$ & \bfslast \\ \midrule
Rate & 79.5\%       & 8.8\%   & 1.8\%       & 2.1\%         \\ \bottomrule \\ \toprule
 Ref.       & \bs   & \dbs & BS+Sample & Sample \\ \midrule
Rate &  5.0\%    &  1.1\%  &  6.4\%  &     0.8\%  \\
\bottomrule
\end{tabular}
\caption{Incomplete rate ($\downarrow$) with different choices of scoring function in best-$k$ search (top) and reference baselines (bottom). $\alpha=0$ stands for sequential log-likelihood without length adjustment; $\alpha=1.0$ represents \bfsmean{}. We show the lowest incomplete rate under various configurations of temporal decay. See Sec.~\ref{sec:discuss-score} for the definition and discussion. }
\label{tab:fail-score}
\end{table}

\subsection{Choice of Scoring Function}
\label{sec:discuss-score}
In this paper, we experimented with two families of scoring functions: length-normalized sequence log-likelihood and a new memoryless greedy score. We studied how the scoring function works in practice. More particularly, we looked into whether some form of scoring function will cause significant incompletion or search failure. 
We test the incompletion rate in a very strict use case: decoding a summary with at most $T=30$ tokens with a total budget $C=bT=300$. If the model does not reach any EOS token before depth of 30, we consider it as a case of incompletion. 
We show the comparison of the incompletion rate in Table~\ref{tab:fail-score}. The length-normalized sequence log-likelihood is formed as $\frac{1}{l^{\alpha}}\sum_{t=0}^{l} \log p_{\theta}(y_t | \mathbf{y_{<t}}, \mathbf{x})$. The original definition of scoring function, $\alpha=0$, is a failure in the context of best-first search. The reason behind is the monotonic relation of the hypothesis score and the length. Since shorter sequences always have higher score, the greedy property of best-first search will hinder the exploration of longer sequences. Although the weight of temporal decay could be increased, it will change the foundation of the algorithm if the decay is overwhelming the hypothesis score. 

\paragraph{Performance} of \bfslast and \bfsmean is overall good across all datasets. We also notice an interesting difference that \bfsmean prioritizes the quality slightly more than \bfslast while \bfslast enjoys more diversity. For example, \bfslast on QuoRef achieves higher distinctness score but a slightly lower ROUGE score. The difference of scoring function will definitely impact the search strategy and we treat it as a handle of controllability for our algorithm.

\begin{figure}[t]
    \centering
    \footnotesize
    \includegraphics[width=0.482\textwidth]{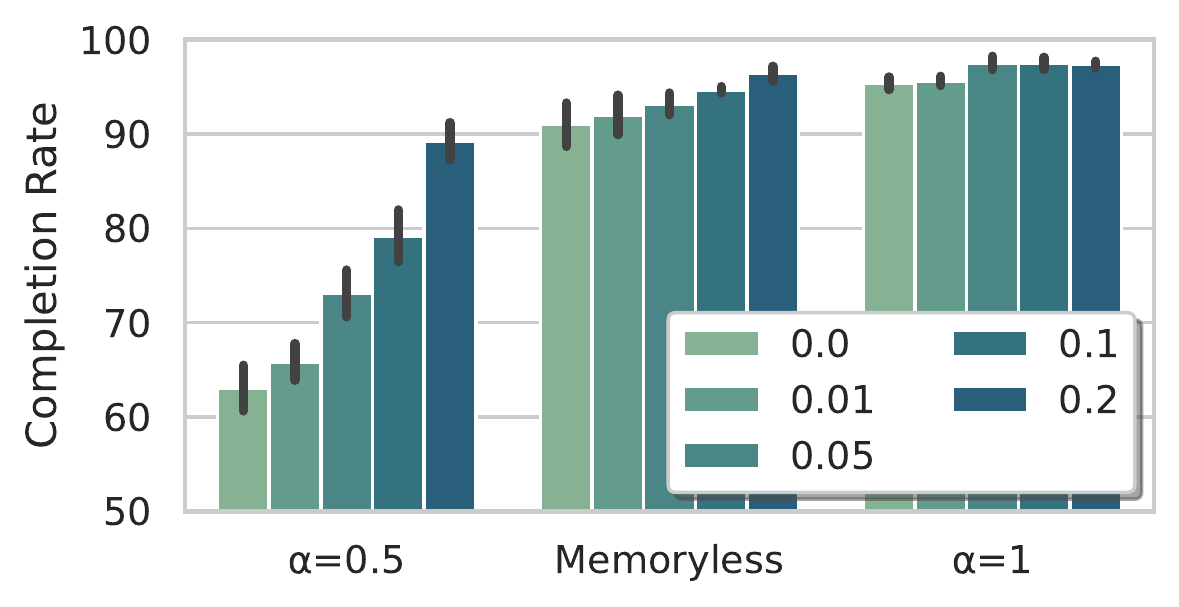}
    \caption{Evaluation of the weight term $\kappa$ for temporal decay. We increase the weight $\kappa$ in the objective from $0.0$ (no decay) to $0.2$ and evaluate how it relates to the completion rate for different scoring functions. In the case of $\alpha=0.5$, increased weight significantly helps the completion rate.  }
    \label{fig:temp}
\end{figure}

\subsection{Effect of Temporal Decay}
We evaluate how temporal decay helps the completion rate in different settings in Figure~\ref{fig:temp}. As the result in Table~\ref{tab:fail-score} indicates a high incomplete rate when $\alpha=0$, we only evaluate three scoring schemas, $\alpha=0.5$, $\alpha=1$ (\bfsmean), and the memoryless setting (\bfslast{}). Temporal decay helps the completion when the scoring function itself struggles with completion. For example, when $\alpha=0.5$, increasing $\kappa$ improves the completion rate from 66\% to 92\%.

\begin{table}[t]
\centering
\footnotesize

\begin{tabular}{@{}r|rcccr@{}}
\toprule
        &          &      & \multicolumn{3}{c}{BLEU} \\ 
        & |S| & \ad   & \textsc{Origin} & \textsc{Comet} & $\Delta$ \\ \midrule
Reference & 11 & 36.9 & - & - & -        \\ \midrule
\bs      & 10      & 15.4 & 30.4 & \textbf{32.3}       & 1.9   \\
\dbs     & 10      & 18.7 & 25.0 & 27.8       & \textbf{2.8}   \\
\dbsplus & 10      & 24.6 & 20.8 & 22.9       & 2.1   \\ \midrule
\btyp{0.2}    & 3      & 11.0 & 26.5 & 26.1       & -0.4  \\
\btyp{0.5}    & 3      & 10.2 & 34.3 & 34.6       & 0.3   \\
\btyp{0.95}    & 3      & 10.7 & 32.9 & 33.4       & \textbf{0.5}   \\
\btopp{0.5}   & 2      & 9.0  & 33.0 & 33.3       & 0.3   \\
\btopp{0.8}   & 3      & 10.2 & 34.9 & \textbf{34.9}       & 0.0   \\
\btopp{0.9}   & 3      & 10.4 & 32.6 & 33.8       & 1.2   \\ \midrule
\typ{0.2}     & 9      & 27.2 & 19.9 & 19.5       & -0.3  \\
\typ{0.5}     & 9      & 28.6 & 25.6 & 27.0       & 1.4   \\
\typ{0.95}     & 10      & 36.5 & 19.2 & 22.1       & \textbf{2.9}   \\
\topp{0.5}    & 8      & 18.6 & 31.1 & \textbf{32.2}       & 1.1   \\
\topp{0.8}    & 10      & 30.2 & 25.9 & 27.0       & 1.0   \\
\topp{0.9}    & 10      & 35.0 & 23.2 & 25.8       & 2.6   \\ \midrule
\bfsmean & 35     & 19.6 & 30.1 & \textbf{33.3}       & 3.2   \\
\bfslast & 33     & 20.5 & 26.1 & 31.1       & \textbf{5.0}   \\ \bottomrule
\end{tabular}
\caption{Machine translation from English to German. \textsc{Origin} and \textsc{Comet} are the BLEU score before and after reranking; $\Delta$ indicates the change of BLEU score from reranking.}
\label{tab:ende}
\end{table}

\section{Application: Reranking Diverse Outputs}
\label{sec:mt}
Machine translation is typically considered as a precision-oriented task, where typically only a few translations are considered as correct. In this section, we will pursue the application of our algorithm in neural machine translation. More specifically, we would like to answer the RQ: \textbf{Do we benefit by selecting from a pool of high-quality diverse outputs, even when the task does not necessarily require such?} 

\paragraph{Setup}
We use a popular machine translation dataset with multiple references \cite{pmlr-v80-ott18a}, based on WMT'14 En-Fr and En-De test sets \cite{bojar-etal-2014-findings}. 
The model for this task is the mBART\footnote{\url{https://huggingface.co/facebook/mbart-large-50-many-to-many-mmt}} model supporting 50 languages \cite{tang-etal-2021-multilingual}. 
In order to rerank decoded outputs, we adopt a state-of-the-art quality estimation model for MT, COMET-QE \cite{rei-etal-2020-comet}. 
The quality estimation model we use to rerank all the outputs is \texttt{wmt21-comet-qe-da}. The QE model is a referenceless model $Q(\mathbf{s}, \mathbf{t})$ which judges whether the source input $\mathbf{s}$ and the hypothesis translation $\mathbf{t}$ form a matched pair based on regression metrics. 

\paragraph{Result} We present the result on MT En-De and En-Fr in Table~\ref{tab:ende} and \ref{tab:enfr}. Our observations include:
\begin{enumerate}
    \item Our approaches have a huge gain after reranking and surpass all of the sampling based methods and beam search only methods by at least 1 point of BLEU.
    \item Compared to other approaches with decent BLEU score, our approaches have high diversity in the outputs. For example, \bs{} and \topp{0.5} are 1.0 and 1.1 BLEU behind \bfsmean{} while they are also 4.2 and 1.0 \ad{} behind. The best BLEU score approach, \btopp{0.8}, is 9.4 from us on \ad{} while the human annotation reference is much higher than any of the machine generated hypothesis sets. 
    \item Diversity-oriented approaches, including pure sampling methods and our methods, generally fall short in BLEU, compared to beam search or beam search based methods. On En-De, beam search based methods obtain up to 34.9 while non beam search based ones get up to 33.3. 
\end{enumerate}
For an increasing number of NLP tasks, users are looking for a diverse set of outputs for various reasons. We show an application here where we can deploy an off-the-shelf reraking model to select a preferred translation from a larger pool. The success of overgeneration-then-reranking paradigm has been witnessed in summarization \cite{song-etal-2021-new,ravaut-etal-2022-summareranker,pernes2022improving} and translation \cite{fernandes-etal-2022-quality}, where the proposed algorithm could be valuable in searching high-quality diverse outputs.

\section{Related Works}

\paragraph{Best-first search} BFS was widely used in structural prediction \cite{klein-manning-2003-parsing}, statistical MT \cite{och-etal-2001-efficient}, and for searching hypotheses \cite{bostrom2022natural,saha2022summarization}. Recent work in decoding strategies \cite{meister-etal-2020-best,lu-etal-2022-neurologic,xu-etal-2022-massive} conceptualized best-first search as part of their paradigm, it was not the dominant component of any of these systems.

\paragraph{Text Decoding Algorithms} Stochastic decoding algorithms have gained popularity in the past few years \cite{fan-etal-2018-hierarchical,Holtzman2020The,meister-typical,suzgun2022crowdsampling}.
Rollout-based algorithms are capable of satisfying certain utility functions or constraints at the cost of efficiency \cite{leblond-etal-2021-machine,chaffin-etal-2022-ppl,lu-etal-2022-neurologic}. 
Recombination-based search algorithm \cite{xu-etal-2022-massive} can find thousands of hypotheses despite complicatedness.

\paragraph{Diversity in Text Generation} The diversity of text generation has been a key challenge for applications like dialogue \cite{li-etal-2016-diversity,zhang-etal-2020-dialogpt,stasaski-hearst-2022-semantic}, MT \cite{shen2019mixture} and conditional text generation \cite{yang-klein-2021-fudge}. 
Beam search has also been developed to generate more diverse outputs \cite{vijayakumar2016diverse,anderson-etal-2017-guided,post-vilar-2018-fast}.
Prior work also studies the trade-off between diversity and quality in text generation \cite{zhang-etal-2021-trading}.

\paragraph{Degeneration of Beam Search} 
\citet{welleck2020neural,Holtzman2020The} addressed the degeneration issue in neural text generation and \citet{pmlr-v97-cohen19a} studies the beam search performance degradation in neural sequence models. 
The gap between high probability and quality has been observed and studied \cite{meister-etal-2022-high,freitag-etal-2022-high}.

\section{Conclusion}
In this work, we propose best-$k$ search, a novel decoding algorithm for text generation based on best-first search. The algorithm features a few technical components, including parallel exploration, temporal decay and heap pruning. The proposed algorithm generates natural and diverse text while maintaining high quality. We conduct comprehensive experiments on four tasks and six datasets to verify the effectiveness of the proposed approach. The algorithm is orthogonal to sampling methods and it is parameter-free, lightweight, efficient, and easy to use.
\section*{Acknowledgements}
We thank Tong Niu, Chen Xing, Hiroaki Hayashi, Katie Stasaski, Philippe Laban, Semih Yavuz and Shafiq Rayhan Joty for helpful careful proofreading and comments on this work. We also thank the rest of the Salesforce AI Research team for generous support and feedback.

\bibliography{anthology,custom}

\appendix

\section{Setup for investigating BFS}
\label{app:init}

We use XSum \cite{narayan-etal-2018-dont} and a BART model \texttt{BART-large-XSum}\footnote{\url{https://huggingface.co/facebook/bart-large-xsum}} \cite{lewis-etal-2020-bart} fine-tuned on it as the testbed of our preliminary study. We sample 100 examples from the test set to measure the decoding quality. We set the beam size to 10 and the max sequence length to 30.
For the machine configuration, we use Intel Xeon CPU @ 2.20GHz for CPU and NVIDIA A100-SXM4-40GB for GPU. 
We use Transformers (v4.23.1) \cite{wolf-etal-2020-transformers} and pytorch (v1.9.0) for baseline implementation and model calls.

\begin{table}[t]
\centering
\footnotesize
\begin{tabular}{@{}r|rcccr@{}}
\toprule
        &          &      & \multicolumn{3}{c}{BLEU} \\ 
        & |S| & \ad   & Original & \textsc{Comet} & $\Delta$ \\ \midrule
Reference & 11 & 29.2 & - & - & -        \\ \midrule
\bs      & 10       & 14.6 & 39.6   & \textbf{38.4}   & -1.2   \\
\dbs     & 10       & 18.4 & 32.1   & 32.1   & 0.0    \\
\dbsplus & 10       & 21.7 & 32.0   & 33.3   & \textbf{1.3}    \\ \midrule
\btyp{0.2}    & 2        & 10.2 & 35.4   & 35.4   & -0.1   \\
\btyp{0.5}    & 2        & 9.2  & 44.3   & \textbf{44.2}   & \textbf{0.0}    \\
\btyp{0.95}    & 3        & 9.9  & 39.9   & 39.7   & -0.2   \\
\btopp{0.5}   & 2        & 8.9  & 40.6   & 40.6   & -0.1   \\
\btopp{0.8}   & 2        & 9.5  & 38.5   & 38.4   & -0.1   \\
\btopp{0.9}   & 3        & 9.8  & 39.5   & 38.9   & -0.6   \\ \midrule
\typ{0.2}     & 8        & 26.4 & 23.9   & 25.0   & 1.1    \\
\typ{0.5}     & 9        & 27.0 & 31.2   & 32.6   & \textbf{ 1.4 }   \\
\typ{0.95}     & 10       & 37.2 & 24.1   & 24.1   & 0.0    \\
\topp{0.5}    & 8        & 17.1 & 35.6   & \textbf{36.3}   & 0.7    \\
\topp{0.8}    & 10       & 28.9 & 28.9   & 28.7   & -0.2   \\
\topp{0.9}    & 10       & 33.4 & 25.4   & 26.6   & 1.2    \\ \midrule
\bfsmean & 18       & 16.8 & 38.0   & \textbf{39.0}   & 1.0    \\
\bfslast & 26       & 18.1 & 33.5   & 37.2   & \textbf{3.6}   \\ \bottomrule
\end{tabular}
\caption{Machine translation from English to French. We highlight the best BLEU score after reranking and the improvement $\Delta$ for each sector. Numbers are rounded after calculation for display simplicity. }
\label{tab:enfr}
\end{table}

\begin{table}[h]
\centering
\footnotesize
\narrowcol
\begin{tabular}{@{}r|rcccccc@{}} 
\toprule
& |S| & \ad  & \ao & \ar & \mv & \grm & \mtr  \\ \midrule
\bs     & 10        & 21.8 & \finecell 55.7 & 41.3 & 91.4          & 87.0         & 48.5           \\
\dbs     & 9         & 25.1 & 50.7 & 36.5 & 72.1          & \badcell 80.9         & 42.1           \\
\dbsplus     & 9         & 29.6 & 50.5 & \badcell 31.9 & \badcell 37.7          & 81.3         & \badcell 35.6           \\ \midrule
\btyp{0.2}     & 1         & 9.8  & \badcell 41.9 & 41.2 & 96.8          & 87.6         & 46.8           \\
\btyp{0.5}     & 1         & 10.2 & 46.1 & \goodcell 44.9 & 94.3          & \goodcell 88.7         & 50.2           \\
\btyp{0.95}     & 2         & 10.7 & 46.6 & \finecell 44.7 & 95.0          & 88.5         & \goodcell 50.4           \\
\btopp{0.5}     & 1         & \badcell 9.2  & 44.8 & 44.5 & 97.0          &  \goodcell 88.7         & 49.6           \\
\btopp{0.8} &1& 10.2&	46.3 &	44.6 & 94.8	& 88.4	&49.9 \\
\btopp{0.9}     & 2         & 10.7 & 46.8 & \finecell 44.7 & 96.0          & 88.1         & \goodcell 50.4           \\
\midrule
\typ{0.2}     & 5         & 21.1 & 45.8 & 37.7 & 94.2          & 87.1         & 43.4           \\
\typ{0.5}     & 7         & 26.2 & 54.3 & 40.6 & 97.1          & 88.1         & 45.7           \\
\typ{0.95}     & 9         & \goodcell 34.8 & 55.6 & 39.2 & 97.8          & 86.5         & 44.6           \\
\topp{0.5}     & 9         & 31.9 & 55.6 & 39.0 & 95.1          & 86.3         & 44.3           \\

\topp{0.8} & 8 & 28.3	&55.4 &	40.8 & 98.4	& 87.8	& 46.1\\
\topp{0.9} & 9 & 31.6	& \goodcell 55.9 &	39.0 & 97.7	& 85.9	&44.3\\
\midrule
\bfsmean  & 19        & 29.5 & 54.8 & 38.2 & \goodcell 99.0          & 86.0         & 44.0           \\
\bfslast  & 18        & \finecell 32.8 & 54.2 & 35.8 & \finecell 98.8          & 84.4         & 40.6          \\ \bottomrule
\end{tabular}
\caption{Results of question generation on SQuAD.}
\label{tab:squad}
\end{table}

\section{Experiment: Machine Translation En$\rightarrow$Fr}
We present the machine translation result from English to French in Table~\ref{tab:enfr}. The dataset we use here is an extended version of newstest2014. We can see a significant improvement over BLEU in our approach after using \textsc{Comet-QE} reranking. 

We obtained similar results on En-Fr compared to En-De in Table~\ref{tab:ende}. Our approach achieves a good combination of diversity and quality compared to baseline methods. One of the beam search + sampling method, \btyp{0.5}, achieves 44.2 after \textsc{Comet-QE} reranking, which surpasses any other methods by a decent margin. Our approach, \bfsmean, beats strong baselines including beam search and sampling-only approaches. What worth noticing is the significant jump after reranking, which shows a great success of overgeneration + reranking as a paradigm.

\section{Experiment: Question Generation on SQuAD}
\label{app:squad}
We present the result of question generation on SQuAD in Table~\ref{tab:squad}. Our approach achieves the best MAUVE score and a good combination of diversity and quality metrics. 

\section{Design Choice for Completion}
\label{app:len}
In our paper, we design a temporal decay function to encourage the completion of our search algorithm.
We have also considered a depth-based auxiliary term to encourage the completion. For instance, we can define $\texttt{aux}(n) = n\texttt{.length()}$ where a longer sequence will receive a higher score if we assume a longer sequence is more likely to terminate \cite{welleck-etal-2020-consistency}. The problem of this function is that it always prefer longer sequences. Once there exists one single long sequence, the rest of the search will focus on this string because it is longer than any other strings. The search will be shaped into a depth-first search while what we expect is to discover a diverse set of strings with various length and prefix.

\end{document}